\documentclass[runningheads]{llncs}
\usepackage[T1]{fontenc}
\usepackage{graphicx}
\usepackage{xcolor}
\usepackage{amsmath}
\usepackage{svg}
\usepackage{amsfonts}
\usepackage{wasysym}
\usepackage{marvosym}
\begin{document}
\title{ANCHOR: Integrating Adversarial Training with Hard-mined Supervised Contrastive Learning for Robust Representation Learning}
\titlerunning{ANCHOR}

\author{Samarup Bhattacharya\inst{1}\textsuperscript{(\(\dagger\))} \and
Anubhab Bhattacharya\inst{1}\textsuperscript{(\(\dagger\))} \and
Abir Chakraborty\inst{1}\textsuperscript{(\(\dagger\))}}
\authorrunning{S. Bhattacharya et al.}
\institute{Department of Computer Science and Engineering, Jadavpur University, Kolkata, India \\
\email{samarupbhattacharjee@gmail.com}\\
\email{anubhabbhattacharya0710@gmail.com}\\
\email{abirc8010@gmail.com}
}
\maketitle            
\renewcommand{\thefootnote}{\fnsymbol{footnote}}
\footnotetext[1]{\textsuperscript{(\(\dagger\))} These authors contributed equally to this work.}
\renewcommand{\thefootnote}{\arabic{footnote}}
\begin{abstract} 
Neural networks have changed the way machines interpret the world. At their core they learn by following gradients, adjusting their parameters step by step until they identify the most discriminant patterns in the data. This process gives them their strength, yet it also opens the door to a hidden flaw. The very gradients that help a model learn can just as easily be used to produce small, imperceptible tweaks that cause the model to completely alter it’s decision. Such tweaks are called adversarial attacks. Adversarial attacks take advantage of this vulnerability by adding tiny, imperceptible changes to images that, while leaving them identical to the human eye cause the model to make wrong predictions. In this work, we propose Adversarially-trained Contrastive Hard-mining for Optimized Robustness (ANCHOR), a framework that leverages the power of supervised contrastive learning with explicit hard positive mining to enable the model to learn representations for images such that the embeddings for the images, their augmentations and their perturbed versions cluster together in the embedding space along with those for the other images of the same class while being separated from the images of other classes. This alignment helps the model focus on stable, meaningful patterns rather than fragile gradient cues. On CIFAR-10, our approach achieves impressive results for both clean and  robust accuracy under PGD-20 ($\epsilon$ = 0.031), outperforming standard adversarial training methods. Our results indicate that combining adversarial guidance with hard-mined contrastive supervision helps models learn more structured and robust representations, narrowing the gap between accuracy and robustness.

\keywords{Adversarial Robustness \and Adversarial Attack \and Supervised Contrastive Learning \and Hard Positive Mining.}
\end{abstract}
\section{Introduction}
Adversarial attacks are intentional, carefully crafted modifications to input data designed to mislead a machine learning model into making incorrect predictions, while keeping the changes almost imperceptible to humans. The concept of adversarial attacks was first introduced by Szegedy et al in 2013 \cite{szegedy2013intriguing}. This was followed up with the 2014 paper by Goodfellow et al \cite{goodfellow2014explaining}, which further framed the problem of adversarial examples and introduced methods (including the fast gradient sign method) for generating them. Later, Kurakin et al \cite{kurakin2018adversarial}, introduced the Iterative FGSM that applies perturbations in multiple small steps. Madry et al \cite{madry2017towards} generalized this into the Projected Gradient Descent (PGD) attack which became a standard benchmark for evaluating model robustness due to its effectiveness and simplicity. Beyond gradient ascent, other attack strategies have also been proposed such as DeepFool by Moosavi-Dezfooli et al. \cite{moosavi2016deepfool} sought minimal perturbations by iteratively linearizing decision boundaries, while Jacobian-based Saliency Map Attack (JSMA) introduced by Papernot et al. \cite{papernot2016limitations} exploited saliency information for targeted misclassification. Afterwards, optimization-based methods such as the Carlini \& Wagner attack \cite{carlini2017towards}  achieved state-of-the-art results, successfully breaking many proposed defenses.

Unlike white-box methods of adversarial attack, black box attacks generate perturbations without access to the gradients of the model. They estimate the optimal perturbations either based on gradient information of another model (surrogate model) or based on estimated gradient information through querying the model. Black-box attacks have been developed to mimic real-world conditions where the internal details of a model such as gradient information are not known to the attacker. Transfer-based black-box attacks are often weaker and less effective than white-box attacks\cite{ilyas2018black} while query/optimization-based attack have generally achieved equal performance albeit at the computational cost of large number of queries\cite{chen2017zoo}. 
Transfer-based approaches \cite{papernot2017practical} first trained local substitute models using label queries and then generated adversarial samples that transferred to the target model. Zeroth-Order Optimization (ZOO) \cite{chen2017zoo} approximated gradients numerically from output scores, eliminating the need for substitute networks. Decision-based attacks such as the Boundary Attack\cite{brendel2017decision} and HopSkipJumpAttack\cite{chen2020hopskipjumpattack} further relaxed assumptions by relying solely on model decisions, making them applicable even in fully restricted query settings. One-Pixel Attack\cite{su2019one} illustrated that even extremely sparse perturbations could suffice for successful misclassification, while query-efficient strategies leveraging natural evolution strategies (NES) and gradient estimation \cite{ilyas2018black} improved the practicality of black-box attacks.

Extensive research has been conducted on improving the robustness of deep learning models against adversarial perturbations. The method we propose builds upon the success of prior works \cite{li2023adversarial}, \cite{zhang2023empowering}, \cite{kim2020adversarial} that have demonstrated the efficacy of contrastive learning approaches in aiding robust performance of models. In particular, we leverage the representational benefits of the supervised contrastive loss introduced by Khosla et al. \cite{khosla2020supervised}. Thus, we make the following contributions:

\begin{enumerate}
    \item We introduce the Adversarially-trained Contrastive Hard-mining for Optimized Robustness framework that uses a supervised contrastive loss paired with adaptive hard positive mining that dynamically assigns greater weights to intra-class samples that are most dissimilar to their class counterparts, ensuring their stronger influence during optimization. This strategy, combined with adversarial training, improves robust accuracy relative to existing baselines.
    \item We evaluate the proposed framework using standard architectures and training settings, enabling direct comparison with established benchmarks.
\end{enumerate}

\begin{figure}
    \centering
    \includegraphics[width=0.8\textwidth]{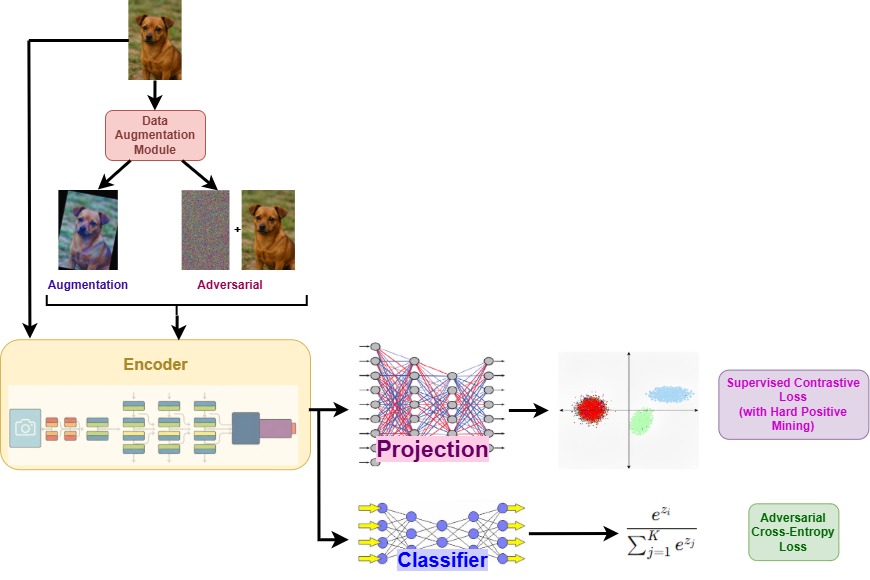}
    \caption{Workflow of the proposed pipeline} 
    \label{fig:Graphical_Abstract}
\end{figure}

\section{Literature Survey}

Representation learning has been central to deep learning research, providing feature spaces that generalize well across tasks. Early unsupervised and self-supervised approaches defined surrogate pretext tasks to learn data-invariant embeddings without labels. Contrastive Predictive Coding (CPC) first modeled representation learning as a prediction task over latent sequences, while contrastive frameworks such as MoCo and SimCLR refined this concept through instance-discrimination objectives. MoCo introduced a momentum encoder and dynamic memory queue to maintain consistent negatives, and SimCLR demonstrated that large-batch contrastive training with strong augmentations can yield highly transferable features. Subsequently, BYOL showed that explicit negatives were unnecessary, achieving comparable results using an online and target-network bootstrap. Together, these approaches established contrastive self-supervision as a dominant paradigm for unsupervised representation learning \cite{oord2018representation}, \cite{he2020momentum},\cite{chen2020simple}, \cite{grill2020bootstrap}.

Supervised contrastive learning extended these ideas by incorporating label information into the contrastive objective. Khosla et al.\ \cite{khosla2020supervised} proposed the Supervised Contrastive Loss, treating all samples from the same class as positives and all others as negatives. This formulation unified supervised and self-supervised objectives, improving both generalization and robustness compared to standard cross-entropy. The resulting representations exhibited tighter intra-class clustering and greater inter-class separability, motivating several subsequent extensions that integrate contrastive supervision into downstream classification and robustness pipelines.

In parallel, the study of adversarial robustness advanced through optimization-based defenses. Projected Gradient Descent (PGD) adversarial training \cite{madry2017towards} remains the empirical benchmark, framing defense as a min–max optimization over perturbations. TRADES \cite{zhang2019theoretically} later introduced a principled regularization to balance natural and robust accuracy, while ensemble adversarial training \cite{tramer2017ensemble} diversified attack sources to mitigate overfitting to specific adversaries. Architectural approaches such as feature denoising \cite{xie2019feature} and adversarial logit pairing \cite{kannan2018adversarial} modified intermediate representations to stabilize model behavior, and randomized smoothing \cite{cohen2019certified} offered certified $\ell_2$ robustness. These methods collectively define the canonical non-representational baselines for evaluating robustness.

Recently, several works have bridged representation learning and robustness. Adversarial Contrastive Learning \cite{li2023adversarial} incorporated adversarial perturbations directly into the contrastive objective, encouraging feature invariance between clean and attacked samples. Zhang et al.\ \cite{zhang2023empowering} demonstrated that adversarially trained contrastive pretraining substantially improves downstream robust accuracy, while Kim et al.\ \cite{kim2020adversarial} analyzed the geometric properties of robust representations, showing tighter intra-class manifolds and larger margins. These findings collectively indicate that robust generalization arises from stable and semantically aligned representations rather than from adversarial augmentation alone.

\section{Proposed Methodology}
In this section, we discuss our proposed \textit{Adversarially-trained Contrastive Hard-mining for Optimized Robustness} in detail. We explain the various components of our model and its overall framework.

\subsection{Dataset Description}
For this study, the standard dataset, namely CIFAR-10, is utilized for model training and inference. The choice of this dataset is motivated by the fact that several benchmark studies involving adversarial robustness of models have utilized and reported their results on this dataset.

\subsection{Overall Framework}
The proposed framework consists of three parts: the data augmentation module, the image encoder, and the multilayer perceptron (MLP) classifier. Following the standard supervised contrastive learning setup \cite{khosla2020supervised}, an additional projection head MLP is present during training, which is discarded during model inference. During model training, the embeddings generated by the image encoder are fed into the projection head and the classifier head. The embeddings from the projection head are used to calculate the supervised contrastive loss, while those from the classifier head are used to calculate the adversarial cross-entropy loss. Finally, a weighted sum is performed over these losses to obtain the training loss for that batch. Later on, a separate MLP classifier is trained using this encoder as a frozen backbone and further adversarial finetuning is performed in accordance to the pipeline utilized by Li et al.\cite{li2023adversarial}.

\subsection{Data Augmentation Module}
In order to learn robust features, the supervised contrastive learning method requires multiple views for each image in the training set \cite{khosla2020supervised}. To this end, our proposed method involves feeding two views for each image: an augmented view $x^{\text{aug}}$ and a perturbed view of the original $x^{\text{adv}}$.  

The perturbed view is generated using Projected Gradient Descent (PGD) attacks:
\vspace{0.5em}
\[
x^{\text{adv}}_{t+1} = \Pi_{x + \mathcal{B}_\epsilon} \big( x^{\text{adv}}_t + \alpha \, \text{sign}(\nabla_x \mathcal{L}_{\text{train}}(f_\theta(x^{\text{adv}}_t), y)) \big), \quad t = 0, \dots, T-1
\]
\vspace{0.5em}
where $\Pi_{x + \mathcal{B}_\epsilon}$ is the projection onto the $\ell_\infty$ ball of radius $\epsilon$, $\alpha$ is the step size, $f_\theta$ is the model, and $\mathcal{L}_{\text{train}}$ is the overall training loss computed as a weighted sum of the supervised contrastive loss with hard positive mining $\mathcal{L}_{\text{SCL}}^{\text{hard}}$ and the adversarial cross-entropy loss $\mathcal{L}_{\text{CE}}^{\text{adv}}$:
\vspace{0.5em}
\[
\mathcal{L}_{\text{train}} = \mathcal{L}_{\text{SCL}}^{\text{hard}} + \lambda \, \mathcal{L}_{\text{CE}}^{\text{adv}},
\]
\vspace{0.5em}
where $\lambda$ is the weighting hyperparameter.
In our experiments, $\epsilon = 0.031$ and $T = 10$ with $\alpha$ = 0.007. Incorporating $x^{\text{adv}}$ during training enables the model to develop representations that are resilient to adversarial perturbations rather than exploiting easily breakable discriminative cues \cite{madry2017towards}.  

The augmented view is generated through a stochastic augmentation function $\mathcal{A}(\cdot)$ applied to the original image:
\vspace{0.5em}
\[
x^{\text{aug}} = \mathcal{A}(x), \quad \mathcal{A} \in \{\text{random crop, flipping, color jittering}\}.
\]
\vspace{0.5em}
This ensures that the model's learned patterns are orientation- and color-information agnostic. These additional views, therefore, promote a robust embedding space where samples of different classes remain well-separated even under input perturbations.

\subsection{Image Encoder}
The image encoder we used was based on the ResNet-18 architecture introduced by He et al. \cite{he2016deep}. While the original ResNet-18 was designed for ImageNet and required 224$\times$224 input images, our setup demanded an encoder capable of handling lower-resolution inputs. To accommodate this, we adopted a modified version of ResNet-18 similar to the design principles in \cite{he2016identity}. Specifically, we replaced the initial 7$\times$7 convolution layer with stride 2 by a 3$\times$3 convolution layer with stride 1 and removed the initial max pooling layer. These adjustments allowed the model to preserve more fine-grained spatial information in the early stages, preventing excessive downsampling and enabling more effective feature extraction from small-scale images.

\subsection{Projection Head}
Several studies have found that a projection head appended to the image encoder, used during model training and discarded during inference helps in convergence for contrastive learning models\cite{gupta2022understanding}, \cite{gui2023unraveling}, \cite{xue2024investigating}. Following these evidences, we kept an MLP projection head that downsampled the embeddings from the image encoder during training and was discarded during model testing. These downsampled embeddings were used for computing the supervised contrastive loss during training.

\subsection{Classifier}
Our choice for a classifier was motivated by the fact that several studies have found that a linear classifier suffices in obtaining an optimum classification performance for the embeddings generated by a contrastive learning framework \cite{chen2020simple}, \cite{chen2021exploring}, \cite{haochen2021provable}, \cite{gupta2022understanding}, \cite{khosla2020supervised}. Additionally, it has been shown that adversarial partial training using an MLP classifier and a frozen backbone after pre-training the backbone using contrastive learning leads to greater performance both in terms of robustness and clean accuracy \cite{li2023adversarial}. Hence, we use an additional MLP classifier which utilizes the frozen backbone pre-trained using the ANCHOR framework as image encoder and learns to classify adversarial samples with a loss function $\mathcal{L}_{\text{CE}}^{\text{adv}}$ defined as
\[
\mathcal{L}_{\text{CE}}^{\text{adv}} = -\frac{1}{N} \sum_{i=1}^{N} \sum_{c=1}^{C} y_{i,c} \log \hat{y}_{i,c}
\]

\subsection{Loss Function}

Following the foundational work of Khosla et al. \cite{khosla2020supervised}, the supervised contrastive loss is widely used to learn class-discriminative yet semantically consistent representations by leveraging label information to identify multiple positive pairs per anchor. The standard supervised contrastive loss is given by:
\[
\mathcal{L}_{\text{SCL}} = \sum_{i=1}^{N} \frac{-1}{|P(i)|} \sum_{p \in P(i)} 
\log \frac{\exp(\mathbf{z}_i \cdot \mathbf{z}_p / \tau)}{\sum_{a=1, a \neq i}^{N} \exp(\mathbf{z}_i \cdot \mathbf{z}_a / \tau)},
\]
where $P(i)$ denotes the set of positive samples belonging to the same class as anchor $i$, and $\tau$ is the temperature parameter. This formulation encourages embeddings from the same class to cluster tightly while separating embeddings from different classes. 

However, treating all positive pairs equally overlooks the fact that not all positive samples contribute equally to representation robustness. In particular, certain positive samples may lie further apart in the embedding space—these \textit{hard positives} capture intra-class diversity and play a critical role in improving robustness and generalization. Prior works has demonstrated that fine-tuning with a diverse set of hard positives improves the robustness networks\cite{rozsa2016adversarial}.

To exploit this observation, we propose the Supervised Contrastive Loss with Hard-Positive Mining, which dynamically emphasizes harder positive samples during training while maintaining stable optimization through curriculum scheduling. Specifically, we introduce a hardness weighting coefficient $\beta$ that evolves smoothly over the course of training:
\[
\beta_t = \beta_{\text{start}} \left( \frac{\beta_{\text{end}}}{\beta_{\text{start}}} \right)^{t / T},
\]
where $t$ is the current epoch and $T$ is the total number of epochs. This exponential scheduling gradually increases the emphasis on hard positives as the model becomes more stable, preventing early training instability.

Given feature representations $\mathbf{z}_i$ and their corresponding labels $y_i$, we define the hardness-adjusted supervised contrastive loss as:
\[
\mathcal{L}_{\text{ANCHOR}} = 
\sum_{i=1}^{N} \frac{-1}{|P(i)|} \sum_{p \in P(i)} 
w_{ip} \cdot \log \frac{\exp(\mathbf{z}_i \cdot \mathbf{z}_p / \tau)}{\sum_{a \neq i}^{N} \exp(\mathbf{z}_i \cdot \mathbf{z}_a / \tau)},
\]
where $w_{ip}$ represents the adaptive weighting factor for each positive pair $(i, p)$, computed inversely to the similarity between anchor and positive embeddings. Thus, harder positives—those less similar to the anchor—receive higher weights, ensuring they exert greater influence on the model’s gradients. This mechanism dynamically adjusts during training through $\beta_t$, allowing the model to progressively focus on informative positives.

In addition to adaptive hard positive mining, we incorporate margin-based handling of negative samples to prevent the over-penalization of easily separable negatives. Together, these mechanisms produce a more discriminative and robust embedding space, particularly effective in adversarially-trained models.

Overall, our proposed loss function balances stable representation learning in early epochs with increased focus on informative and challenging samples later in training, thereby yielding improved adversarial robustness and better structured latent representations.

\section{Results and Discussion}
In order to obtain results directly comparable to existing benchmarks, we used model architecture and training settings as laid out by authors of existing benchmarks \cite{li2023adversarial} and evaluated our framework on the CIFAR-10 dataset. As explained in a previous section, we used ResNet-18 as the image encoder with batch size 512, a projection head used during training to downsize the embeddings before calculating the contrastive loss and discarded during model testing and a linear classifier to calculate adversarial cross-entropy loss. Together, they constituted the pipeline that was pre-trained using our proposed ANCHOR Loss. After pre-training, this pipeline was frozen and the image encoder was used along with a zero-initialized MLP classifier to for further adversarial partial training(APT). APT involved freezing the image encoder and finetuning the MLP classifier through adversarial training using attacks with the same specifications to generate the adversarial samples as were used for the pre-training i.e. PGD-10 attacks with $\epsilon$ = 0.031, $\alpha$ = 0.007 for training. We evaluated our model against white-box attacks since, in general, they are stronger and more efficient than black-box attacks and usually, a greater robust accuracy against white-box attacks directly implies an even greater performance against black-box ones. For model testing, PGD-20 attacks with $\epsilon$ = 0.031, $\alpha$ = 0.003 were used along with AutoAttacks. Robust accuracies against PGD attacks and AutoAttack(RA-PGD and RA-AA respectively) as well as clean accuracy(CA) are reported in Table \ref{tab:ci_results} and compared against prior works on robustness.

\begin{table}[hbt]
\centering
\caption{Performance comparison of baseline robustness methods and the proposed framework on CIFAR-10 image classification. Since, our training settings match with those detailed in \cite{li2023adversarial}, for comparison, we use the baseline results as reported by Li et al. under conditions of standard partial finetuning(SPF), adversarial partial training(APT) and adversarial full finetuning(AFF). Reported metrics include RA-PGD(\%), RA-AA(\%) and CA(\%)}
\label{tab:ci_results}
\begin{tabular}{|c|c|c|c|}
\hline
Method & RA-PGD & RA-AA & CA \\
\hline
SimCLR(SPF) & 0.27 & 0.00 & 90.60 \\
RoCL(SPF) & 40.27 & 28.38 & 83.71 \\
Selfie + DPE(AFF) & 52.22 & 40.24 & 83.00 \\
ACL(AFF) & 52.82 & 45.61 & 82.19 \\
AdvCL(APT) & 52.01 & 43.52.66 & 79.39 \\
AT & 44.05 & 40.07 & 84.48 \\
TRADES & 51.41 & 45.41 & 82.20 \\
SCL & 0.11 & 0.00 & \textbf{92.01} \\
ASCL(APT) & 53.09 & 45.7 & 81.67 \\
\hline
ANCHOR(APT) & \textbf{54.10} & \textbf{46.07} & 81.40 \\
\hline
\end{tabular}
\end{table}

As shown in Table \ref{tab:ci_results}, our proposed ANCHOR framework achieves the highest robust accuracies among all compared methods, with RA-PGD of 54.10\% and RA-AA of 46.07\%, while maintaining a competitive clean accuracy of 81.40\%. This performance improvement over existing contrastive and adversarial training approaches such as AdvCL, ASCL, and TRADES demonstrates the efficacy of integrating hard-mined supervised contrastive learning with adversarial objectives during pre-training. The results indicate that the ANCHOR Loss promotes the learning of more discriminative and robust representations that generalize better to adversarial perturbations. Furthermore, the improvement under the APT setting validates the observation by Li et al. \cite{li2023adversarial} that after pre-training, adversarial finetuning of the MLP classifier is enough to obtain high performance eliminating the need for the computationally expensive fine-tuning of the encoder backbone.

\section{Conclusion}
In this work, we proposed ANCHOR, a novel framework that unifies adversarial training with hard-positive-mined supervised contrastive learning to enhance model robustness and representation quality. Through comprehensive evaluation on CIFAR-10, ANCHOR consistently outperforms existing benchmarks in robust accuracy while preserving competitive clean accuracy, validating the effectiveness of our approach. The results highlight the potential of leveraging hard positive samples during contrastive learning to improve adversarial resilience. Future work could extend ANCHOR to other datasets and explore its adaptability across different architectures and attack types.
\begin{credits}
\subsubsection{\ackname} This research received no external funding.

\subsubsection{\discintname}
Authors report no conflict of interest.
\end{credits}
\bibliographystyle{splncs04}
\bibliography{references}
\end{document}